\documentclass[10pt,twocolumn,letterpaper]{article}

\usepackage[accsupp]{axessibility} % Improves PDF readability for those with disabilities.
\usepackage{iccv}              % To produce the CAMERA-READY version
\usepackage{multirow}
\usepackage{colortbl}
\usepackage{pifont}
\usepackage{amssymb}
\definecolor{iccvblue}{rgb}{0.21,0.49,0.74}
\usepackage[pagebackref,breaklinks,colorlinks,allcolors=iccvblue]{hyperref}

\def\paperID{4972} % *** Enter the Paper ID here
\def\confName{ICCV}
\def\confYear{2025}

\title{Free-MoRef: Instantly Multiplexing Context Perception Capabilities of Video-MLLMs within Single Inference}

\author{Kuo Wang$^{1,2,4}$\thanks{Equally-contributed authors.}\quad
Quanlong Zheng$^{3*}$\quad
Junlin Xie$^{5}$\quad
Yanhao Zhang$^{3}$\thanks{Project Leader.}\quad 
Jinguo Luo$^{6}$\quad 
Haonan Lu$^{3}$\quad 
\\
Liang Lin$^{1,2,8}$\quad
Fan Zhou$^{1,4,7}$\quad
Guanbin Li$^{1,2,4,8}$\thanks{Corresponding author.}\quad\\
{$^1$Sun Yat-sen University, $^2$Peng Cheng Laboratory, $^3$OPPO AI Center, OPPO Inc., China}  \\
$^{4}$Research Institute, Sun Yat-sen University, Shenzhen, China \\
$^{5}$The Chinese University of Hong Kong, Shenzhen, China \\
$^{6}$Harbin Institute of Technology, Shenzhen, China \\
$^{7}$Shenzhen Key Laboratory of Digital Living Network and Content Service \\
$^{8}$Guangdong Key Laboratory of Big Data Analysis and Processing \\
{\tt\small wangk229@mail2.sysu.edu.cn, \{zhengquanlong,zhangyanhao,luhaonan\}@oppo.com} \\
{\tt\small junlinxie@link.cuhk.edu.cn, 23s153135@stu.hit.edu.cn }\\
{\tt\small linliang@ieee.org, \{isszf,liguanbin\}@mail.sysu.edu.cn}
}

\begin{document}
\maketitle
\begin{abstract}
Video Multimodal Large Language Models~(Video-MLLM) have achieved remarkable advancements in video understanding tasks. However, constrained by the context length limitation in the underlying LLMs, existing Video-MLLMs typically exhibit suboptimal performance on long video scenarios. To understand extended input frames, common solutions span token compression and streaming inference techniques, which sacrifice feature granularity or inference efficiency. Differently, to efficiently achieve comprehensive understanding of longer frame inputs, we draw ideas from MoE and propose a training-free approach \textbf{Free-MoRef}, which instantly multiplexes the context perception capabilities of Video-MLLMs within one inference pass. Specifically, Free-MoRef reconstructs the vision tokens into several short sequences as multi-references. Subsequently, we introduce MoRef-attention, which gathers clues from the multi-reference chunks in parallel to summarize unified query activations. After the shadow layers in LLMs, a reference fusion step is derived to compose a final mixed reasoning sequence with key tokens from parallel chunks, which compensates the cross-reference vision interactions that are neglected in MoRef-attention. By splitting and fusing the long vision token sequences, Free-MoRef achieves improved performance under much lower computing costs in reasoning multiplexed context length, demonstrating strong efficiency and effectiveness. Experiments on VideoMME, MLVU, LongVideoBench show that Free-MoRef achieves full perception of 2$\times$ to 8$\times$ longer input frames without compression on a single A100 GPU while keeping instant responses, thereby bringing significant performance gains, even surpassing dedicatedly trained long-video-MLLMs. Codes are available at \url{https://github.com/wkfdb/Free-MoRef}
\end{abstract}    
\section{Introduction}
\label{sec:intro}

\begin{figure}[t]
  \centering
  \begin{subfigure}{\linewidth}
    \includegraphics[width=\linewidth]{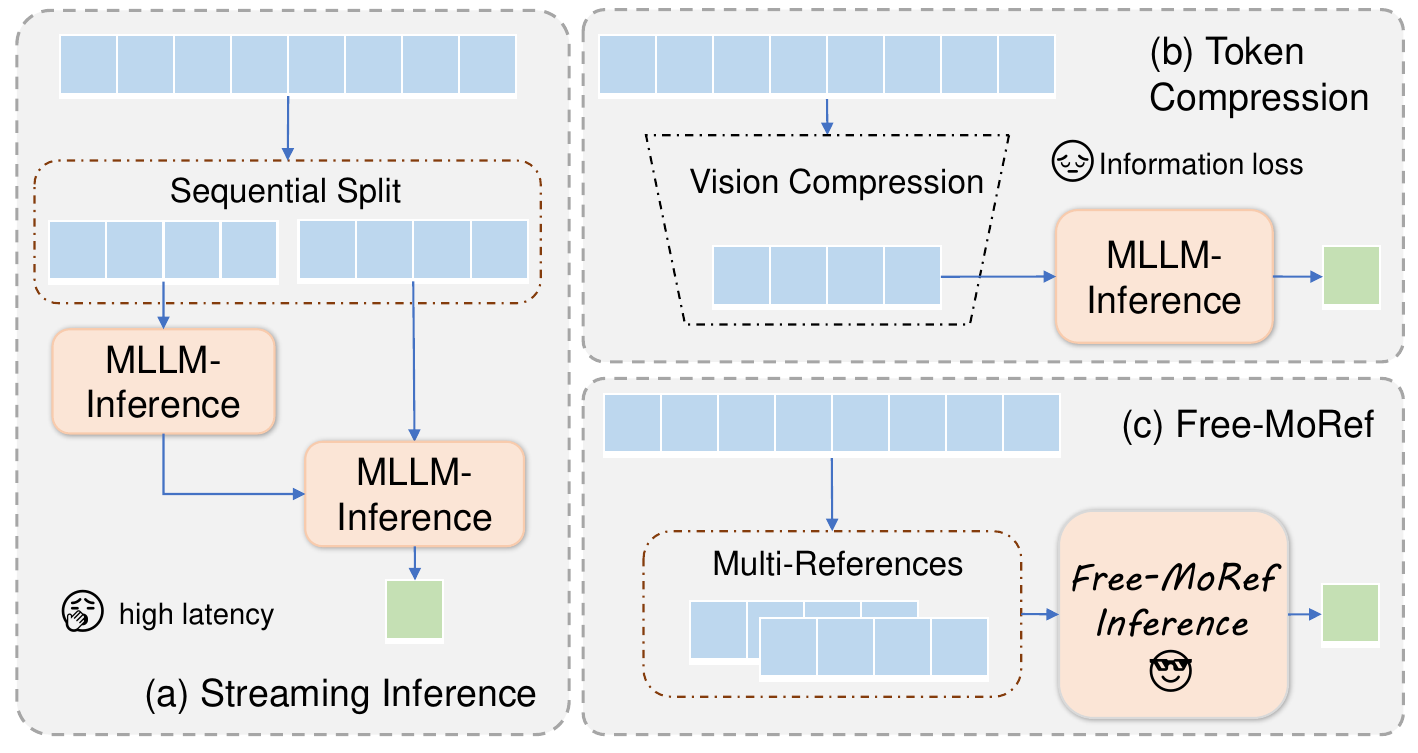}
    \caption{Different designs to expand the context perception capability. Free-MoRef achieves both efficient and comprehensive understanding of the multiplexed vision inputs.}
    \label{fig1a}
  \end{subfigure}
  \begin{subfigure}{\linewidth}
    \includegraphics[width=\linewidth]{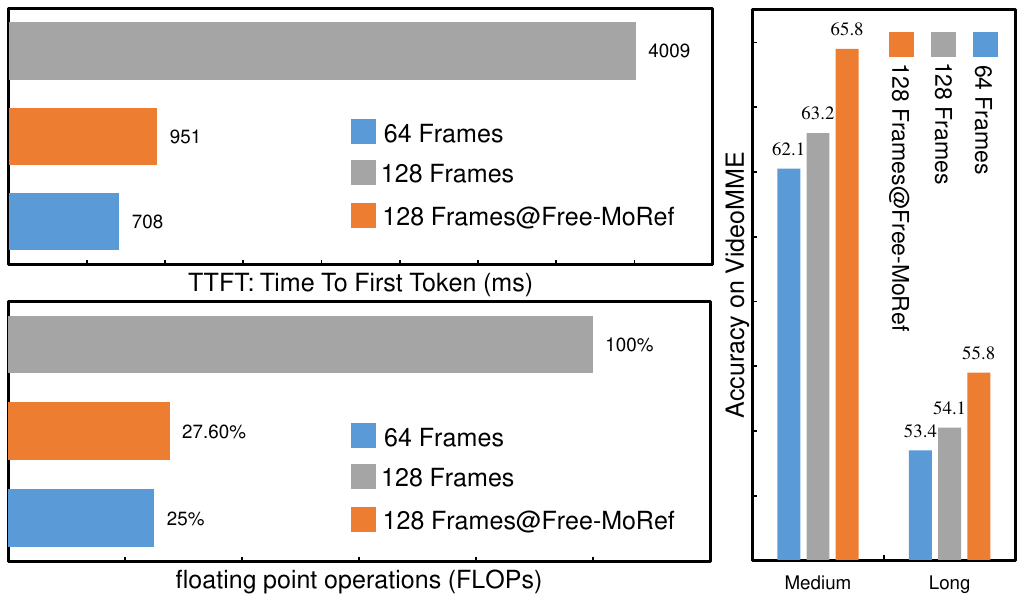}
    \caption{Comparison of FLOPs, first token latency and overall QA accuracy in reasoning original and doubled input frames by LLaVA-Video~\cite{zhang2024video}.}
    \label{fig1b}
  \end{subfigure}
  \caption{Different inference designs and the advantages of Free-MoRef. In summary, Free-MoRef brings superior performance under much lower computing costs on longer vision contexts.}
  \label{fig1}
%\vspace{-15pt}
\end{figure}

Large Language Models (LLMs)~\cite{achiam2023gpt, yang2024qwen2, touvron2023llama}have emerged as a revolutionary force towards general intelligence, marked by their universal capabilities in various language tasks. Through instruction tuning~\cite{liu2023visual}, Multimodal Large Language Models~(MLLMs) further extend their exceptional reasoning ability to other modalities such as vision~\cite{wang2024qwen2, li2024llava, team2024internvl2} and audio~\cite{monfort2021spoken}. In recent studies, MLLMs have been extensively applied to the comprehension of video content~\cite{maaz2024videogpt+, chen2025sharegpt4video, li2023videochat, wang2024internvideo2}. Notwithstanding the remarkable advances in video understanding, the context length restriction inherent in LLMs has emerged as a critical bottleneck, especially for long video understanding, where the abundant visual tokens readily surpass the threshold that ensures stability and consequently resulting in a decline in the effectiveness of these models.

To prevent over-length sequences that violate the context constraints of the reasoning LLM, existing MLLMs~\cite{wang2024qwen2, zhang2024llava, team2024internvl2} typically impose a maximum limit on the length of vision tokens to maintain stable performance. (e.g. 64 frames with $2\times2$ spatial pooling for LLaVA-Video~\cite{zhang2024video} and 784 small rescaled frames for Qwen2-VL~\cite{wang2024qwen2}.) The tradeoff between resolution and number of input frames have to be made in video understanding models, which greatly restricts their effectiveness in fully exploiting the rich information contained within extended video sequences.

To expand the context perception capability under the context restriction of the foundation LLM in Video-MLLMs, common solutions primarily encompass token compression~\cite{shu2024video, ren2023testa, li2024llama} and streaming inference technique~\cite{xiao2023efficient, ning2024inf, wang2024retake}. However, both of these methods suffer from notable deficiencies. The Streaming Inference technique~\cite{xiao2023efficient} achieves ultra-long context dependency by retaining and reusing the historical KV-CACHE, but the extra time cost is proportional to the context length benefit. For example, reasoning doubled contexts results in doubled latency. As an alternative, the token compression strategy can represent more information within a limited token length, thereby increasing the context within a single inference without exceeding the preset token length limit. However, longer context benefits lead to more severe information loss. In light of these drawbacks, a crucial question emerges: \textit{Is it possible to achieve longer context perception within a single inference while ensuring comprehensive understanding of the context?}

Motivated by this question, we have designed and implemented Free-MoRef, a training-free approach that instantly multiplexes the context throughput within one single inference pass, achieving full long-context understanding with flash efficiency. Inspired by the MoE paradigm~\cite{dai2024deepseekmoe}, we abstract long visual tokens into multiple short sequences as multiple references, each of which encapsulates the overall information of the original long contexts. Subsequently, we further design the \textbf{M}ixture \textbf{o}f \textbf{Ref}erence attention, which allows for the parallel querying of multiple references and the integration of the results into a unified activation within each decoding layer. This process could be considered as an expert solving problems according to different references and figuring out a final solution. As observed in FastV~\cite{chen2024image}, after the shadow layers in LLM, the attention pattern would be more concentrated on query tokens. Leveraging this insight, we further extract key vision tokens in each chunk and mix them into a global reference for the remaining decoding layers, which compensates the neglected cross-reference vision interactions in the parallel reasoning. As illustrated in Figure~\ref{fig1b}, by splitting and fusing the vision-tokens, Free-MoRef achieves instant comprehensive understanding of longer contexts with improved performance, demonstrating strong efficiency and effectiveness. 

We apply Free-MoRef to LLaVA-Video-7B~\cite{zhang2024video} and conduct experiments on several long video benchmarks, including VideoMME~\cite{fu2024video}, MLVU~\cite{zhou2024mlvu}, and LongVideoBench~\cite{wu2025longvideobench}. On a single A100 GPU, Free-MoRef can directly extend the context throughput from 2$\times$ to 8$\times$ with less than 27.6\% of the FLOPs and negligible latency, while bringing superior performance on all the three above long video benchmarks. On the VideoMME benchmark, our method leads 3\% to 5\% performance gains on long and medium videos, reaching SOTA results, even surpassing dedicatedly trained long-video-MLLMs~\cite{shu2024video,xue2024longvila,zhang2024long}.  Notably, Free-MoRef supports Flash-Attention~\cite{dao2022flashattention} and can also be integrated with streaming inference or token compression strategies. Despite training-free application, the MoRef-attention mechanism may also inspire the training design for long context scenarios.

\section{Related Works}
\label{sec:related-works}

%-------------------------------------------------------------------------
\subsection{Video Large Language Models}

Recent advancements in Video-MLLMs are mainly achieved by empowering Image-MLLM to comprehend video content. Here, large-scale video-text data is utilized to learn the vision features and temporal relations between the input frames. The structure of Video-MLLM typically consists of a vision-encoder to tokenize the vision inputs, a foundation LLM for query-aware reasoning, and an intermediate connector to link the vision space and language space. In terms of the connector, several works~\cite{li2023videochat, cheng2024videollama,li2024mvbench} use the Q-Former~\cite{li2023blip} to merge visual and text features, where learnable tokens summaries the encoded patch embeddings. However, high compression rate usually results in lower performance in such designs. In other approaches~\cite{lin2023video, zhang2024llava, maaz2023video, ataallah2024minigpt4}, patch embeddings are directly concatenated, which is more effecitve since it preserves more detailed features. However, the abundance of vision tokens poses a vital challenge in understanding longer videos.

\begin{figure*}[t]
  \centering
  \includegraphics[width=\linewidth]{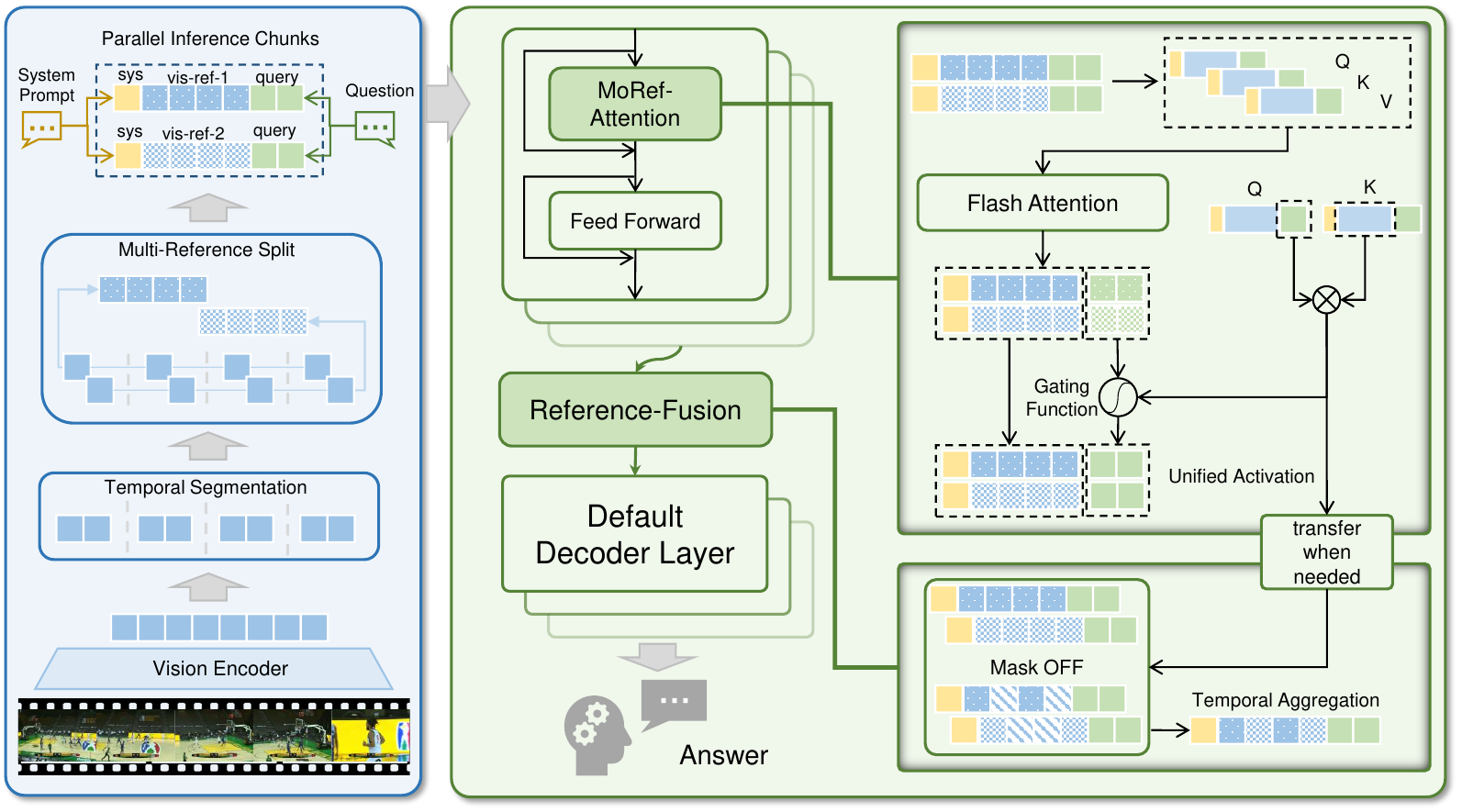}
  \caption{The framework of Free-MoRef Inference. For extended input frames, the initial step involves partitioning the vision tokens into multiple references and subsequently assigning identical system prompt and question to each of these references. To enable efficient comprehension of these multi-reference chunks, we design the MoRef attention mechanism, which concurrently extracts clues from multiple references to formulate responses to the posed question. At the middle layer of the decoder, a reference-fusion step is derived. This step serves to aggregate the parallel chunks into a unified global representation which not only further accelerates the reasoning process but also facilitates cross-chunk interactions, thereby enhancing the overall performance and effectiveness of the long context reasoning.}
  \label{fig2}
% \vspace{-15pt}
\end{figure*}

\subsection{Long Video Understanding}
Through uniform sampling, existing Video-MLLMs can be directly applied to long video understanding tasks. However, it is evident that the limited input severely restricts the model's performance. To tackle the challenge of long video understanding, existing research can be mainly categorized into three types: LLM context expansion~\cite{team2024gemini, wei2024visual, xue2024longvila, fei2024video}, token compression~\cite{shu2024video, li2024llama, weng2024longvlm, ren2023testa}, and streaming inference~\cite{ning2024inf, wang2024retake}.
Context extension methods aim to directly increase the context length limit of Video-MLLMs by conducting post-training with long-sequence data, thereby enhancing the model's context perception ability. Although these methods are effective, they impose a substantial computational burden on long video understanding, which restricts their application in practical scenarios.
Regarding vision token compression, training-free methods~\cite{huang2024prunevid, chen2024image, shang2024llava} prune visual tokens based on spatiotemporal redundancy, while training-based approaches~\cite{shu2024video, chen2024fewer} introduce learnable summary tokens to achieve token compression. However, a low compression rate may not yield significant contextual improvement, while a high compression rate could result in substantial information loss.
Recently, streaming inference techniques~\cite{xiao2023efficient} have been applied to long video understanding~\cite{wang2024retake}. This is achieved by invoking Video-MLLM multiple times to gradually comprehend the long context through identifying and reusing key historical KV-CACHE. Nevertheless, this process leads to exponential reasoning delays.

Our approach bears resemblance to the context extension method. However, the key distinction lies in the fact that our method is training-free, incurs low computational overhead, and can instantly achieve comprehensive perception of the exponentially increased context within less than 27.6\% of the computational cost.

%-------------------------------------------------------------------------

%-------------------------------------------------------------------------

\section{Method}
In this section, we present the Free-MoRef method, which effectively extends the context perception capacity of Video-MLLMs with a high degree of flexibility. Notably, when dealing with extended input frames, Free-MoRef initially partitions the long vision tokens into parallel inference chunks via multi-reference partitioning. Subsequently, it substitutes the self-attention layers of the LLM with MoRef attention. This allows for parallel reasoning over multiple references using the same query and aggregation of unified activations. At the mid-decoder layers, an optional Reference Fusion step is derived to combine the parallel chunks, thereby further enhancing the efficiency of the reasoning process. Without additional training, Free-MoRef overcomes the context length constraint of Video-MLLM during single inference and attains comprehensive perception of exponentially expanding context with minimal computational cost and achieves better performance. The overall architecture of Free-MoRef Inference is depicted in Figure 2, and the details of each component will be elucidated in the following subsections.

\subsection{Multi-Reference Partition}

In Video-MLLM, the reasoning LLM typically has a context length threshold to safeguard stable performance. For instance, the LLaVA-Video~\cite{zhang2024video} model employs Qwen2~\cite{yang2024qwen2technicalreport} as its underlying LLM, and the corresponding sequence length threshold is set at 32768 tokens. When the length of the inference sequence surpasses this threshold, it frequently results in performance deterioration or, more critically, Out-Of-Memory errors. Under these constraints, existing approaches commonly resort to vision token compression or streaming inference techniques to handle longer input frames. To minimize information loss while concurrently ensuring the efficiency of the reasoning process, we propose to partitioning the long visual token sequence into multiple parallel chunks, which serve as multi-references for comprehensive understanding.

To enhance flexibility, we initially divide the vision-token sequence into $M$ units according to the temporal relationship. Subsequently, within each unit, we further temporally decompose it into $N$ fragments. Here, both $M$ and $N$ are manually configured hyperparameters. Eventually, through the aggregation of fragments from diverse units, we are able to obtain $N$ reference chunks. Each of these references can be considered as an abstraction of the extended video sequence. Notably, the larger the value of the hyperparameter $M$, the more pronounced the temporal intersection among the references. When setting $M=1$, the $N$ chunks will be temporally independent of each other. After completing the multi-reference partition, we assign identical system prompt and question to each vision sequence, thereby forming the final parallel inferecne chunks, which is designed for more comprehensive and efficient reasoning by leveraging the \textbf{M}ixture-\textbf{o}f-\textbf{Ref}erence (MoRef) attention. 

\subsection{MoRef Attention}
MoRef Attention is the key step in attaining comprehensive perception and parallel reasoning across multi-references. It concurrently queries distinct references in parallel using an identical question and combines multiple attention outcomes to summarize a unified response for updating the question tokens within each decoder layer.

Specifically, for the input parallel inference chunks, MoRef first constructs $\boldsymbol{Q}, \boldsymbol{K}, \boldsymbol{V} \in \mathbb{R}^{N\times l\times D}$ ($N$ is the chunk number, $l$ is the sequence length and $D$ is the embedding dimension), then executes flash-attention to obtain the initial attention results $\boldsymbol{O}$, where $\boldsymbol{O} = [\boldsymbol{O}^{sys}, \boldsymbol{O}^{vis}, \boldsymbol{O}^{ques}]$. Here, $\boldsymbol{O}^{sys}$, $\boldsymbol{O}^{vis}$ and $\boldsymbol{O}^{ques}$ respectively denote the attention results corresponding to the system prompt token, vision token, and question token. Owing to the unidirectional nature of causal attention, $\boldsymbol{O}^{sys}$ in different chunks would be exactly the same. In contrast, $\boldsymbol{O}^{vis}$ and $\boldsymbol{O}^{ques}$ yield divergent results due to the variance in vision-references. At this point, we maintain the variation of $\boldsymbol{O}^{vis}$ and aggregate $\boldsymbol{O}^{ques}$ across different chunks through the following function: 
\begin{equation}
\boldsymbol{O}^{fusion} = (\sum_{i=1}^{N}\omega_i\cdot\boldsymbol{O}^{ques}_{i}).repeat(N),\\ \sum_{i=1}^{N}\omega_i=1
\label{eq1}
\end{equation}

In Eq.~\ref{eq1}, $\boldsymbol{O}^{ques}_{i}$ represents the query result on each reference and $\boldsymbol{O}^{fusion}$ is the unified summarization. By replacing $\boldsymbol{O}^{ques}$ in the initial attention results, the output of MoRef attention is constructed as $\boldsymbol{O}^{MoRef} = [\boldsymbol{O}^{sys}, \boldsymbol{O}^{vis}, \boldsymbol{O}^{fusion}]$.

The $\omega_i$ in Eq~\ref{eq1} indicates a gating function, which controls the information aggregation across different references. The gating function should be query-aware, as the key information required to answer the question may not be uniformly distributed among diverse references. In the training-free implementation, our objective is to seek the query-reference-correlation from the attention map. Since the flash-attention doesn't support output attention weight, we manually calculate the multi-model attention map between query and vision-reference:
\begin{equation}
\boldsymbol{A} = softmax(\boldsymbol{Q}^{ques}\times (\boldsymbol{K}^{vis})^T )
\label{eq2}
\end{equation}

Compared with the full-attention on the whole sequence, the cross-modal attention Eq.~\ref{eq2} introduces negligible computation.  With respect to $\boldsymbol{A}$, we set the gating weights of each reference chunk as:
\begin{equation}
\omega_i = \frac{max(\boldsymbol{A}[i])}{\sum_{i=1}^{N}max(\boldsymbol{A}[i])}
\label{eq3}
\end{equation}

\begin{figure}[t]
  \centering
  \includegraphics[width=\linewidth]{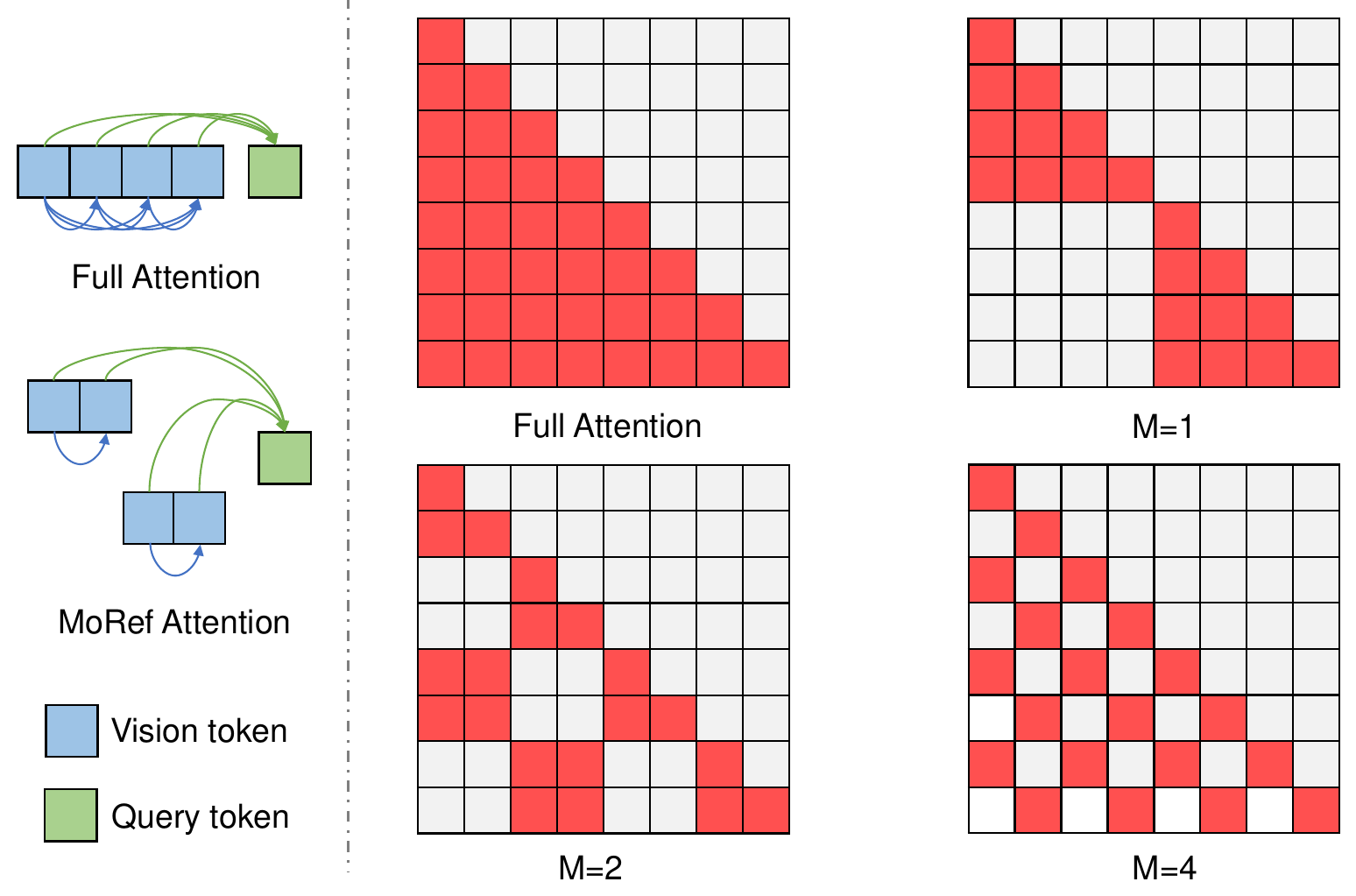}
  \caption{Visualization of attention patterns of full attention and MoRef attention. MoRef attention achieves the full-awareness of vision tokens, while the attention map among vision tokens are sparsified. The right part represents vision attention map of splitting 8 vision tokens into two chunks under different setting of temporal unit number ($M$).
}
  \label{fig3}
%\vspace{-15pt}
\end{figure}

By combining the attention results of the same query across diverse vision-references, all vision-tokens are effectively engaged in the updating of the query-token in each decoder layer. As shown in Figure~\ref{fig3}, this integration strategy enables the full-context perception equivalent to that achieved by full attention. By partitioning the vision sequence into $N$ non-overlapping chunks, the computational complexity is reduced by approximately a factor of $\frac{1}{N}$ compared with full attention. The number of temporal units $M$ serves as a crucial parameter that significantly influences the formation of sparse attention patterns. For instance, consider a vision sequence of length 8 divided into $N = 2$ chunks. Figure~\ref{fig3} depicted the impact of different settings of $M$ on the vision-attention-map.

\subsection{Reference Fusion}
MoRef attention efficiently enables the comprehensive perception of multi-references. Nevertheless, the division of multi-references disrupts the connectivity among vision-tokens across different chunks. To address this limitation, we design an additional Reference Fusion step. This step aims to achieve the integration of multiple reference chunks into a global one, thereby compensating for the lack of cross-chunk interaction within the deep decoder layer.

The implementation of Reference Fusion is grounded in an observation made by FastV~\cite{chen2024image}: vision-tokens contribute uniformly in the shallow decoder layers. In contrast, within the deep layers, the attention weights of the decoder layer would more concentrate on the query-token. We visualized the reasoning process of LLaVA-Video~\cite{zhang2024video} and noted a similar phenomenon~(as detailed in supplementary material). Leveraging this insight, we maintain parallel multi-reference reasoning within the shallow layer. When the decoding process reaches a specific layer $L$, we perform the merging of multi-references based on the attention map $\boldsymbol{A}$ computed in Eq~\ref{eq2}.

Specifically, $\boldsymbol{A}\in\mathbb{R}^{N\times l_{ques}\times l_{vis}}$, where $N$ represents the number of references, $l_{ques}$ and $l_{vis}$ denote the number of question token and vision token in each inference chunk. We compute the average of the attention map $\boldsymbol{A}$ along the $l_{ques}$ dimension to construct the importance estimation matrix $\boldsymbol{E}\in\mathbb{R}^{N\times l_{vis}}$, where each element $E_{ij}$ in $\boldsymbol{E}$ quantifies the average contribution of the $j$-th vision token in the $i$-th chunk. Based on the estimation matrix $\boldsymbol{E}$, we prune $1-\frac{1}{N}$ of the less important vision-tokens within each inference chunk. Subsequently, we aggregate the remaining vision tokens into a global reference in accordance with their temporal relationships. System prompt tokens and question tokens are directly transferred from the local reference chunk to the global reference. For the following decoding process, only the global reference is used by the default decoder layers of the LLM.

Through the Reference fusion step, the pruning of non-crucial tokens further reduces the computational load, while the cross-chunk vision interaction that is lacking in shallow layers can be effectively compensated for, which results in optimized performance. %Our experiments demonstrate that in a double-chunk reasoning scenario, when the reference fusion is executed at the third layer, the computational cost can be further reduced to 27.6\%, while achieving superior performance.

\section{Experiments}
\subsection{Experimental Setup}
\paragraph{Benchmarks}
\textbf{VideoMME.}~\cite{fu2024video} Video Multi-Modal Evaluation benchmark (VideoMME) consists of 900 videos with a total duration of 256 hours, covering a wide range of video types. The videos are associated with 2,700 manually labeled complex multiple-choice QA pairs across 30 subfields. According to video durations, VideoMME is partitioned into three subsets : short ($<$ 2 minutes), medium (4 $\sim$ 15 minutes), and long (30 $\sim$ 60 minutes). \textbf{MLVU.}~\cite{zhou2024mlvu} Multi-Task Long Video Understanding Benchmark (MLVU) significantly expands the scope of durations with diverse types of videos and 7 different types of QA tasks. The video lengths range from 3 minutes to over 2 hours, with an average duration of 12 minutes. \textbf{LongVideoBench.}~\cite{wu2025longvideobench} LongVideoBench highlights referred reasoning questions, which are dependent on long frame inputs. It contains 17 finer-grained question categories on 10 different types of videos. The video duration covers 4 groups: 8-15 seconds, 15-60 seconds, 3-10 minutes, and 15-60 minutes.

\begin{table*}[h]
\renewcommand\tabcolsep{6pt}
\begin{center}
\newcommand{\tabincell}[2]{\begin{tabular}{@{}#1@{}}#2\end{tabular}}  %导言区
  \caption{Performance of Free-MoRef@LLaVA-Video-7B under extended frame inputs. The red color indicates Out-Of-Memory error on a single A100 GPU. We managed the inference under the help of \textit{accelerate} toolkit.}
  \begin{tabular}{lcccccccc}
  \toprule
  \multirow{2}*{\tabincell{l}{\textbf{Context Length}\\({Token Number})}}  & \multirow{2}*{\textbf{FLOPs}} & \multirow{2}*{\textbf{MLVU}} & \multicolumn{3}{c}{\textbf{VideoMME}} & \multicolumn{3}{c}{\textbf{LongVideoBench}} \\
  \cline{4-9}
   & & & \textbf{Medium} & \textbf{Long} & \textbf{Overall} & \textbf{600s} & \textbf{3600s} & \textbf{Overall}\\
   \midrule
   64 frames (11648) & 100\% & 70.3 & 62.1 & 53.4 & 64.3 & 60.4 & 51.2 & 58.8\\
   \midrule
   128 frames (23296) & 400\% & 70.2 & 63.2 & 54.1 & 64.9 & 60.6 & 50.8 & 58.7\\
   \rowcolor{blue!5}128 frames@Free-MoRef & \textbf{110.4\%} & \textbf{70.8} & \textbf{65.8} & \textbf{55.8} &\textbf{66.3} & \textbf{62.1} & \textbf{51.0} & \textbf{59.3}\\
   \midrule
   \textcolor{red}{256 frames (46592)} & 1600\% & 67.2 & 61.4 & 54.1 & 63.1 & 57.2 & 48.5 & 56.7\\
   \rowcolor{blue!5}256 frames@Free-MoRef & \textbf{163.2\%} & \textbf{72.5} & \textbf{66.4} & \textbf{55.3} &\textbf{66.3} & \textbf{62.1} & \textbf{51.2} & \textbf{59.3}\\
   \midrule
   \textcolor{red}{512 frames (93184)} & 6400\% & 61.1 & 55.7 & 48.8 & 60.6 & 53.1 & 45.9 & 54.3\\
   \rowcolor{blue!5}512 frames@Free-MoRef & \textbf{400\%} & \textbf{72.8} & \textbf{67.3} & \textbf{56.0} &\textbf{66.9} & \textbf{62.8} & \textbf{51.9} & \textbf{59.9}\\

  %\multicolumn{8}{c}{\underline{MLVU}} & \multicolumn{15}{c}{\underline{VideoMME}} & \multirow{2}*{Latency} \\
     %& AO & AO & AO & AO & AO & AO & AO & Avg & AO & AO & AO & AO & AO & AO & AO & AO & AO & AO & AO & AO & Med & Long & Avg &  \\
	%\multirow{2}*{\tabincell{c}{Models}} & \multicolumn{2}{c|}{\underline{VL-PLM}} & \multicolumn{2}{c}{\underline{MarvelOVD}}\\
	% & $AP_{50}^{N}$ & $AP_{50}^{B}$ & $AP_{50}^{N}$ & $AP_{50}^{B}$\\
	%\midrule
	%$\gamma=1$ & 32.7  &   54.0  &   37.8   &   57.0 \\
	%$\gamma=2$ & 32.5  &   53.9  &  \textbf{38.9}   &   56.5 \\
	%$\gamma=4$ & -  &   -  &   38.6   &   56.0 \\
  \bottomrule
  \end{tabular}
  \label{tab1}
\end{center}
%\vspace{-15pt}
\end{table*}

\begin{table}[h]
\renewcommand\tabcolsep{2pt}
\begin{center}
\newcommand{\tabincell}[2]{\begin{tabular}{@{}#1@{}}#2\end{tabular}}  %导言区
  \caption{Performance comparison on Long-Video Benchmarks: all models in this table are of the 7B $\sim$ 8B scale.}
  \begin{tabular}{lcccc}
  \toprule
  \multirow{2}*{\textbf{Method}} & \multirow{2}*{\textbf{MLVU}} & \multirow{2}*{\textbf{LVideoBench}} & \multicolumn{2}{c}{\textbf{VideoMME}} \\
  \cline{4-5}
   & & & \textbf{Long} & \textbf{Overall} \\
   \midrule
   InternVL2~\cite{team2024internvl2} & 64.0 & 54.6 & - & 54.0\\
   InternVL2.5~\cite{chen2024expanding} & 68.4 & 57.5 & 53.0 & 64.5\\
   Qwen2-VL~\cite{wang2024qwen2} & 64.8 & 55.6 & 55.7 & 63.3\\
   \tabincell{l}{LLaVA-\\OneVision}~\cite{li2024llava} & 64.7 & 56.3 & - & 58.2\\
   LLaVA-Video~\cite{zhang2024video} & 70.2 & 58.2 & 53.4 & 64.3\\
   Kangaroo~\cite{liu2024kangaroo} & 61.0 & 54.8 & 46.7 & 56.0\\
   LongVILA~\cite{xue2024longvila} & - & 57.1 & 47.0 & 60.1\\
   LongVA~\cite{zhang2024long} & 56.3 & - & 46.2 & 52.6\\
   Video-XL~\cite{shu2024video} & 64.9 & 50.7 & - & 55.5\\
   RETAKE~\cite{wang2024retake} & 69.8 & - & 56.2 & 63.9 \\
   \rowcolor{blue!5}\tabincell{l}{LLaVA-Video\\@Free-MoRef} & \textbf{72.8} & \textbf{59.9} & \underline{56.0} & \textbf{66.9}\\
  \bottomrule
  \end{tabular}
  \label{tab2}
\end{center}
%\vspace{-15pt}
\end{table}

\paragraph{Implementation Details}
We implement the Free-MoRef on the LLaVA-Video-7B~\cite{zhang2024video} model. 
%which is composed of SigLip~\cite{zhai2023sigmoid} as the vision encoder and Qwen2~\cite{yang2024qwen2technicalreport} as the reasoning LLM. 
%The input frames are resized to a resolution of 448×448 pixels, and each frame is represented by 182 tokens. 
By default, LLaVA-Video-7B loads videos at a FPS=1, with a maximum of 64 frames, where each frame is represented by 182 tokens. To demonstrate the efficacy of Free-MoRef, we multiply the maximum number of frames to 128 (2x), 256 (4x), and 512 (8x) respectively.
In our experimental setup, the vision token sequence is consistently partitioned into $M = 64$ temporal units. For the multiplexed inputs of 2x, 4x, and 8x, the number of parallel chunks is set as $N = 2,4,8$ respectively, and the reference fusion layer is configured as $L = 3,6,12$. We conduct the evaluation upon \textit{lmms-eval} framework~\cite{zhang2024lmmsevalrealitycheckevaluation} and all experiments are executed on a single A100 GPU. Further details will be made accessible in our publicly-released code repository.
%, facilitating reproducibility and in-depth analysis by the research community.

\subsection{Main Results}

\begin{table*}[ht]
\renewcommand\tabcolsep{6pt}
\begin{center}
\newcommand{\tabincell}[2]{\begin{tabular}{@{}#1@{}}#2\end{tabular}}  %导言区
  \caption{Performance comparison on various task categories in VideoMME. Tasks contain Temporal Perception(TP), Spatial Perception(SP), Attribute Perception(AP), Action Recognition(ARec), Object Recognition(ORec), OCR Problems(OCR), Counting Problem(CP), Temporal Reasoning(TR), Spatial Reasoning(SR), Action Reasoning(AR), Object Reasoning(OR) and Information Synopsis(IS). The best result is \textbf{bolded}, the second is \underline{underlined}, and the worst is in \textcolor{red}{red}.}
  \begin{tabular}{lccccccccccccc}
  \toprule
  \textbf{Context Length} & TP & SP & AP & ARec & ORec & OCR & CP & TR & SR & AR & OR & IS & Avg\\
   \midrule
   64 frames & \underline{74.5} & \textbf{64.8} & \textbf{79.3} & \textcolor{red}{64.5} & \underline{71.5} & \textcolor{red}{66.9} & \textbf{48.5} & \textcolor{red}{48.0} & \textcolor{red}{80.4} & 56.1 & \textcolor{red}{59.0} & \textcolor{red}{76.5} & \textcolor{red}{64.3}\\
   \midrule
   128 frames & \textcolor{red}{70.9} & \textcolor{red}{59.3} & \textbf{79.3} & \underline{66.8} & \textcolor{red}{70.6} & \underline{71.2} & \textcolor{red}{47.4} & \textcolor{red}{48.0} & \textcolor{red}{80.4} & \textcolor{red}{54.7} & \underline{61.2} & 78.9 & \underline{64.9} \\
   
   \tabincell{l}{128 frames\\@Free-MoRef} & \underline{74.5} & \textbf{64.8} & \underline{78.8} & \textbf{67.1} & \textbf{72.3} & \textbf{71.9} & \textbf{48.5} & \textbf{50.8} & \textbf{85.7} & \underline{58.2} & \textbf{61.9} & \underline{79.6} & \textbf{66.3} \\
   \midrule
   \tabincell{l}{256 frames\\@Free-MoRef} & \textbf{80.0} & \underline{63.0} & \textcolor{red}{77.9} & \textbf{67.1} & \textbf{72.3} & 70.5 & \underline{47.8} & \underline{50.3} & \underline{82.1} & \textbf{60.0} & \textbf{61.9} & \textbf{80.8} & \textbf{66.3} \\
   
  \bottomrule
  \end{tabular}
  \label{tab3}
\end{center}
%\vspace{-15pt}
\end{table*}

\paragraph{Results on Multiplexed Context Understanding.} We implemented the Free-MoRef method on the llava-video-7B model, which by default has a maximum input frame number of 64. In order to verify Free-MoRef's ability to handle multiplexed contexts, we expanded the maximum number of input frames to 128, 256, and 512 for experiments. The experimental results are presented in Table~\ref{tab1}. 

When the number of input frames is doubled, the number of tokens required to encode the input frames amounts to 23,296, which is within the context length limit of Qwen2 (32,768). Under such circumstances, the performance of the model remains nearly unchanged on both the MLVU and LongVideoBench, while demonstrating a slight improvement on the VideoMME benchmark. After the application of Free-MoRef, the computational cost incurred during the inference process is reduced by (1-110.4/400 = 72.4\%). Concurrently, both MLVU and LongVideoBench exhibit a 0.5\% performance gain. The performance improvement on VideoMME is more pronounced, with a 2.6\% improvement for medium-length videos and a 1.7\% improvement for long videos.

When the number of input frames is quadrupled, the number of vision tokens surpasses 40,000. This length clearly exceeds the context length limit of Qwen2, and attempting to perform inference using a single A100 GPU will directly result in an Out-Of-Memory error. Notably, even without the assistance of the \textit{accelerate} toolkit, our proposed method is capable of effectively reasoning on up to 512 frames using a single A100 GPU. 

Leveraging the \textit{accelerate} toolkit, we were enabled to conduct further comparative experiments. When the number of input frames reached 256, the model's performance experienced a substantial decline across all benchmarks. However, upon the application of Free-MoRef, the degradation was effectively mitigated. In particular, on the MLVU dataset, which consists of relatively longer videos, the model's performance was further enhanced to 72.5\%. In terms of efficiency, the computational requirement for the model to infer a context of 4$\times$ length was merely 163.2\% of the original. When the number of input frames is further increased by 8$\times$, reaching 512 frames, the length of the vision token approaches nearly 100,000. In comparison to the scenario with 256-frame input, the performance of the baseline model experiences a substantial and further decline. Despite posing challenges to the baseline model, the extended context serves as a rich source of information and provides more abundant references for the reasoning process of Free-MoRef, as a result, Free-MoRef attains a further enhancement in performance. In summary, Free-MoRef enables the Video-MLLM to leverage the multiplexed frame inputs for more comprehensive understanding of long videos, thereby highlighting the robustness and efficiency of Free-MoRef in handling ultra-long context scenarios.

\paragraph{Comparison with other Models.} 
We validated the efficacy of Free-MoRef by conducting comprehensive comparisons with other Video-MLLMs. These comparisons encompassed open-source MLLMs capable of video understanding, as well as specifically designed long-video understanding models. As depicted in Table~\ref{tab2}, our proposed Free-MoRef method outperformed all the others, attaining the optimal results on the MLVU, LongVideoBench and VideoMME benchmarks.

The underlying reason for the SOTA performance lies in the fact that Free-MoRef enables an efficient and exhaustive understanding of ultra-long contexts. By simply expanding the input frames, Free-MoRef can achieve superior long-video understanding performance within a single inference. It is worth noting that our method is implemented in a training-free manner, which further confirms the potential of MoRef-attention. Its ability to fully understanding ultra-long contexts while maintaining a low computational burden offers significant inspiration for the development of future long-video understanding models, thereby highlighting the practical value and far-reaching implications of Mixture-of-Reference design in the field of long video understanding tasks.

\section{Ablation \& Analysis}
In this section, we perform ablation experiments and in-depth analysis on the Free-MoRef method. Free-MoRef is principally associated with three hyperparameters: the number of sequential units $M$, the number of reference partitions $N$, and the specific decoder layer $L$ for reference fusion. We conduct detailed ablations based on 128-frame inputs on the VideoMME benchmark to evaluate the impact of the hyperparameters. Please refer to supplementary materials for additional analysis.

\subsection{Perforamance on different types of question.}
Table~\ref{tab3} records the performance of varying context inputs across different types of questions. In general, compared with 64-frame input, the expansion of the context predominantly brings benefits in Information Synopsis, and diverse Recognition and Reasoning questions. By applying Free-MoRef for long-context inference, the performance of nearly all types of question on the VideoMME benchmark are enhanced, except for Attribute Perception task. Here is an example of an AP task. \textit{Question-id 009-1: Which color of clothes is QuYuan wearing in the video?} This type of question only refers to a small clip in the video. Expanding the context introduces redundant unnecessary information in this case, thus impact the accuracy of answering such questions.

\begin{table}[t]
\renewcommand\tabcolsep{4pt}
\begin{center}
\newcommand{\tabincell}[2]{\begin{tabular}{@{}#1@{}}#2\end{tabular}}  %导言区
  \caption{Effects of key components of Free-MoRef.}
  \begin{tabular}{ccc|c}
  \toprule
  \tabincell{c}{\textbf{Multi-Reference}\\ \textbf{Partition}} & \tabincell{c}{\textbf{MoRef}\\ \textbf{Attention}} & \tabincell{c}{\textbf{Reference}\\ \textbf{Fusion}}  & \textbf{Overall}\\
   \midrule
   \ding{55} & \ding{55} & \ding{55} & 64.9\\
   \ding{55} & \ding{55} & \checkmark & 63.9\\
   \checkmark & \ding{55} & \checkmark & 62.0\\
   \checkmark & \checkmark & \ding{55} & \underline{65.8}\\
   \midrule
   \checkmark & \checkmark & \checkmark & \textbf{66.3}\\
  \bottomrule
  \end{tabular}
  \label{tab4}
\end{center}
%\vspace{-15pt}
\end{table}
\subsection{Effects of each components.}
In Table~\ref{tab4}, we perform ablation experiments on the key components of Free-MoRef. Directly applying Reference Fusion at the third layer without Multi-Reference Partition and MoRef-Attention is equivalent to dropping 50\% of the vision tokens using the FastV~\cite{chen2024image} method, which inevitably results in a performance decline. Building upon this baseline, applying Multi-Reference Partition to reconstruct the input vision sequence into two chunks and conducting inference with full attention separately leads to a further deterioration in performance. However, when MoRef-Attention is utilized to fuse the attention results across multiple reference, a significant improvement is observed. This clearly demonstrates that Free-MoRef enhances the contextual understanding capabilities of Video-MLLM primarily through the parallel reasoning of MoRef-Attention over Multi-References. Moreover, implementing Reference Fusion on the foundation of MoRef-Attention can further optimize the performance. This indicates that establishing connections among the vision references of different chunks could further help the overall understanding.

\begin{table}[t]
\renewcommand\tabcolsep{4pt}
\begin{center}
\newcommand{\tabincell}[2]{\begin{tabular}{@{}#1@{}}#2\end{tabular}}  %导言区
  \caption{Performance of different setting of parallel chunk number $N$. $N=1$ indicates default inference.}
  \begin{tabular}{ccccc}
  \toprule
  \textbf{Chunk Number} & \textbf{FLOPs} & \textbf{Medium} & \textbf{Long} & \textbf{Overall}\\
   \midrule
   $N=1$ & 100\% & 63.2 & 54.1 & 64.9 \\
   \midrule
   \rowcolor{blue!5}$N=2$ & 27.6\% & 65.8 & 55.8 & 66.3 \\
   $N=4$ & 25\% & 65.1 & 55.8 & 66.1 \\
   $N=8$ & 23.6\% & 64.9 & 55.6 & 65.9 \\
   
  \bottomrule
  \end{tabular}
  \label{tab5}
\end{center}
%\vspace{-15pt}
\end{table}

\subsection{Impacts of various reference number N.}
The Free-MoRef method partitions the video sequence into $N$ identically sized parallel references. In general, the more parallel references there are, the more computational effort is saved during inference. In the context of a 128-frame input, we set $M = 128 / N$ to keep the vision-tokens are split frame-by-frame into each reference chunk. The effects of the model under different values of $N$ are presented in Table~\ref{tab5}. As the number of reference chunk increases, both the computational load and the performance of Free-MoRef exhibit a gradual decline. For best performance, we identically set $N =$ \textit{input frame number} $/ 64$ for all the experiments. 

\subsection{Effects of different temporal units M.}
\begin{table}[t]
\renewcommand\tabcolsep{3pt}
\begin{center}
\newcommand{\tabincell}[2]{\begin{tabular}{@{}#1@{}}#2\end{tabular}}  %导言区
  \caption{Performance of different setting of temporal units $M$. TP indicates Temporal Perception task and SP represents Spatial Perception.}
  \begin{tabular}{cc|cc|c}
  \toprule
  \tabincell{c}{\textbf{Temporal} \textbf{Units}}  &\tabincell{c}{\textbf{Partition} \textbf{Units}} & \textbf{TP} & \textbf{SP} & \textbf{Overall}\\
   \midrule
   $M=1$ & 64 frames & 70.9 & 68.5 & 66.3 \\
   $M=4$ & 16 frames & 72.7 & 66.7 & 66.0 \\
   $M=32$ & 4 frames & 74.5 & 66.7 & 66.0 \\
   $M=64$ & 1 frames & 74.5 & 64.8 & 66.3 \\
  \bottomrule
  \end{tabular}
  \label{tab6}
\end{center}
%\vspace{-15pt}
\end{table}

The configuration of the temporal unit \(M\) determines the temporal intersection among different reference chunks. Its influence on the vision attention pattern is illustrated in Figure~\ref{fig3}. Table~\ref{tab6} documents the effects of diverse values of \(M\) on various tasks. Overall, the performance remains relatively consistent. However, significant disparities are observed in Temporal Perception and Spatial Perception tasks. We provide examples of TP and SP questions for better explanation. \textit{TP: In which part of the video does the red parrot appear?} \textit{SP: What is the location of the scene being depicted in the video?} When \(M = 1\) and $N=2$, the middle part of the entire video serves as the tail and head for the first and second reference chunks, which leads to opposed interpretations of the temporal perception task, thereby reducing the performance for TP tasks. In the case of spatial perception problems, a detailed comprehension of a continuous video segment is essential. When \(M = 64\), the specific video segments are separated to each reference chunk frame by frame, which reduces the feature density and thereby impacting the reasoning for SP questions.

\subsection{Analysis of reference fusion layer L.}

\begin{table}[t]
\renewcommand\tabcolsep{6pt}
\begin{center}
\newcommand{\tabincell}[2]{\begin{tabular}{@{}#1@{}}#2\end{tabular}}  %导言区
  \caption{Performance of different Reference Fusion Layers $L$.}
  \begin{tabular}{c|c|ccc}
  \toprule
  \textbf{Context} & \tabincell{c}{\textbf{Fusion}} & \textbf{Medium} & \textbf{Long} & \textbf{Overall}\\
   \midrule
   \multirow{3}*{\tabincell{c}{128 frames\\ drop rate\\ 50\%}} & \ding{55} & 64.7 & 55.4 & 65.8 \\
    & $L=1$ & 64.4 & 54.4 & 65.4 \\
    & $L=3$ & 65.8 & 55.8 & 66.3 \\
    % & $L=4$ & 64.8 & 55.3 & 65.7 \\
    \midrule
    \multirow{3}*{\tabincell{c}{256 frames\\ drop rate\\ 75\%}} & \ding{55} & 64.6 & 54.8 & 65.5 \\
    & $L=3$ & 64.4 & 56.3 & 66.0 \\
    & $L=6$ & 66.4 & 55.3 & 66.3 \\
  \bottomrule
  \end{tabular}
  \label{tab7}
\end{center}
%\vspace{-15pt}
\end{table}

By discarding unimportant vision tokens in the middle layer and merging multi-reference chunks, further computational savings can be realized while the missing visual feature dependencies between parallel chunks can be made up. However, it is crucial to note that premature execution of reference fusion may lead to a certain degree of information loss, which impairs the final performance.
As depicted in Table~\ref{tab7}, when conducting inference on a context of 128-frame length, performing fusion operations at the first layer yields sub-optimal performance. In the case of reasoning about a length of 256 frames, executing fusion operations at the third layer results in superior long video understanding performance compared to performing it at the sixth layer. This phenomenon indicates that the timely establishment of visual feature associations across parallel chunks is more conducive to longer video understanding. From a holistic perspective, performing reference fusion at the sixth layer can achieve relatively higher performance at the overall evaluation.

\section{Conclusion}
% \section{Limitation and Future Works}

In this paper, we present Free-MoRef, a novel training-free approach that instantly multiplexes the context perception capacity of VideoLLM within a single inference pass. By partitioning long video inputs into multi-reference chunks, our proposed MoRef-attention concurrently extracts clues from multi-references and synthesize unified query responses, thus facilitates the understanding of long videos with extended input frames. Leveraging Free-MoRef, we successfully achieved a comprehensive understanding of 1024 frames using a 7B-VideoLLM on a single A100 GPU and brought substantial improvements across three long video understanding benchmarks. 
{
    \small
    \bibliographystyle{ieeenat_fullname}
    \bibliography{main}

\begin{thebibliography}{45}
\providecommand{\natexlab}[1]{#1}
\providecommand{\url}[1]{\texttt{#1}}
\expandafter\ifx\csname urlstyle\endcsname\relax
  \providecommand{\doi}[1]{doi: #1}\else
  \providecommand{\doi}{doi: \begingroup \urlstyle{rm}\Url}\fi

\bibitem[yan(2024)]{yang2024qwen2technicalreport}
Qwen2 technical report, 2024.

\bibitem[Achiam et~al.(2023)Achiam, Adler, Agarwal, Ahmad, Akkaya, Aleman, Almeida, Altenschmidt, Altman, Anadkat, et~al.]{achiam2023gpt}
Josh Achiam, Steven Adler, Sandhini Agarwal, Lama Ahmad, Ilge Akkaya, Florencia~Leoni Aleman, Diogo Almeida, Janko Altenschmidt, Sam Altman, Shyamal Anadkat, et~al.
\newblock Gpt-4 technical report.
\newblock \emph{arXiv preprint arXiv:2303.08774}, 2023.

\bibitem[Ataallah et~al.(2024)Ataallah, Shen, Abdelrahman, Sleiman, Zhu, Ding, and Elhoseiny]{ataallah2024minigpt4}
Kirolos Ataallah, Xiaoqian Shen, Eslam Abdelrahman, Essam Sleiman, Deyao Zhu, Jian Ding, and Mohamed Elhoseiny.
\newblock Minigpt4-video: Advancing multimodal llms for video understanding with interleaved visual-textual tokens.
\newblock \emph{arXiv preprint arXiv:2404.03413}, 2024.

\bibitem[Chen et~al.(2024{\natexlab{a}})Chen, Zhao, Liu, Bai, Lin, Zhou, and Chang]{chen2024image}
Liang Chen, Haozhe Zhao, Tianyu Liu, Shuai Bai, Junyang Lin, Chang Zhou, and Baobao Chang.
\newblock An image is worth 1/2 tokens after layer 2: Plug-and-play inference acceleration for large vision-language models.
\newblock In \emph{European Conference on Computer Vision}, pages 19--35. Springer, 2024{\natexlab{a}}.

\bibitem[Chen et~al.(2025)Chen, Wei, Li, Dong, Zhang, Zang, Chen, Duan, Tang, Yuan, et~al.]{chen2025sharegpt4video}
Lin Chen, Xilin Wei, Jinsong Li, Xiaoyi Dong, Pan Zhang, Yuhang Zang, Zehui Chen, Haodong Duan, Zhenyu Tang, Li Yuan, et~al.
\newblock Sharegpt4video: Improving video understanding and generation with better captions.
\newblock \emph{Advances in Neural Information Processing Systems}, 37:\penalty0 19472--19495, 2025.

\bibitem[Chen et~al.(2024{\natexlab{b}})Chen, Yuan, Chen, Jie, and Ma]{chen2024fewer}
Shimin Chen, Yitian Yuan, Shaoxiang Chen, Zequn Jie, and Lin Ma.
\newblock Fewer tokens and fewer videos: Extending video understanding abilities in large vision-language models.
\newblock \emph{arXiv preprint arXiv:2406.08024}, 2024{\natexlab{b}}.

\bibitem[Chen et~al.(2024{\natexlab{c}})Chen, Wang, Cao, Liu, Gao, Cui, Zhu, Ye, Tian, Liu, et~al.]{chen2024expanding}
Zhe Chen, Weiyun Wang, Yue Cao, Yangzhou Liu, Zhangwei Gao, Erfei Cui, Jinguo Zhu, Shenglong Ye, Hao Tian, Zhaoyang Liu, et~al.
\newblock Expanding performance boundaries of open-source multimodal models with model, data, and test-time scaling.
\newblock \emph{arXiv preprint arXiv:2412.05271}, 2024{\natexlab{c}}.

\bibitem[Cheng et~al.(2024)Cheng, Leng, Zhang, Xin, Li, Chen, Zhu, Zhang, Luo, Zhao, et~al.]{cheng2024videollama}
Zesen Cheng, Sicong Leng, Hang Zhang, Yifei Xin, Xin Li, Guanzheng Chen, Yongxin Zhu, Wenqi Zhang, Ziyang Luo, Deli Zhao, et~al.
\newblock Videollama 2: Advancing spatial-temporal modeling and audio understanding in video-llms.
\newblock \emph{arXiv preprint arXiv:2406.07476}, 2024.

\bibitem[Dai et~al.(2024)Dai, Deng, Zhao, Xu, Gao, Chen, Li, Zeng, Yu, Wu, et~al.]{dai2024deepseekmoe}
Damai Dai, Chengqi Deng, Chenggang Zhao, RX Xu, Huazuo Gao, Deli Chen, Jiashi Li, Wangding Zeng, Xingkai Yu, Yu Wu, et~al.
\newblock Deepseekmoe: Towards ultimate expert specialization in mixture-of-experts language models.
\newblock \emph{arXiv preprint arXiv:2401.06066}, 2024.

\bibitem[Dao et~al.(2022)Dao, Fu, Ermon, Rudra, and R{\'e}]{dao2022flashattention}
Tri Dao, Dan Fu, Stefano Ermon, Atri Rudra, and Christopher R{\'e}.
\newblock Flashattention: Fast and memory-efficient exact attention with io-awareness.
\newblock \emph{Advances in neural information processing systems}, 35:\penalty0 16344--16359, 2022.

\bibitem[Fei et~al.(2024)Fei, Li, Deng, Wang, Liu, and Wang]{fei2024video}
Jiajun Fei, Dian Li, Zhidong Deng, Zekun Wang, Gang Liu, and Hui Wang.
\newblock Video-ccam: Enhancing video-language understanding with causal cross-attention masks for short and long videos.
\newblock \emph{arXiv preprint arXiv:2408.14023}, 2024.

\bibitem[Fu et~al.(2024)Fu, Dai, Luo, Li, Ren, Zhang, Wang, Zhou, Shen, Zhang, et~al.]{fu2024video}
Chaoyou Fu, Yuhan Dai, Yongdong Luo, Lei Li, Shuhuai Ren, Renrui Zhang, Zihan Wang, Chenyu Zhou, Yunhang Shen, Mengdan Zhang, et~al.
\newblock Video-mme: The first-ever comprehensive evaluation benchmark of multi-modal llms in video analysis.
\newblock \emph{arXiv preprint arXiv:2405.21075}, 2024.

\bibitem[Huang et~al.(2024)Huang, Zhou, and Han]{huang2024prunevid}
Xiaohu Huang, Hao Zhou, and Kai Han.
\newblock Prunevid: Visual token pruning for efficient video large language models.
\newblock \emph{arXiv preprint arXiv:2412.16117}, 2024.

\bibitem[Li et~al.(2024{\natexlab{a}})Li, Zhang, Guo, Zhang, Li, Zhang, Zhang, Zhang, Li, Liu, et~al.]{li2024llava}
Bo Li, Yuanhan Zhang, Dong Guo, Renrui Zhang, Feng Li, Hao Zhang, Kaichen Zhang, Peiyuan Zhang, Yanwei Li, Ziwei Liu, et~al.
\newblock Llava-onevision: Easy visual task transfer.
\newblock \emph{arXiv preprint arXiv:2408.03326}, 2024{\natexlab{a}}.

\bibitem[Li et~al.(2023{\natexlab{a}})Li, Li, Savarese, and Hoi]{li2023blip}
Junnan Li, Dongxu Li, Silvio Savarese, and Steven Hoi.
\newblock Blip-2: Bootstrapping language-image pre-training with frozen image encoders and large language models.
\newblock In \emph{International conference on machine learning}, pages 19730--19742. PMLR, 2023{\natexlab{a}}.

\bibitem[Li et~al.(2023{\natexlab{b}})Li, He, Wang, Li, Wang, Luo, Wang, Wang, and Qiao]{li2023videochat}
KunChang Li, Yinan He, Yi Wang, Yizhuo Li, Wenhai Wang, Ping Luo, Yali Wang, Limin Wang, and Yu Qiao.
\newblock Videochat: Chat-centric video understanding.
\newblock \emph{arXiv preprint arXiv:2305.06355}, 2023{\natexlab{b}}.

\bibitem[Li et~al.(2024{\natexlab{b}})Li, Wang, He, Li, Wang, Liu, Wang, Xu, Chen, Luo, et~al.]{li2024mvbench}
Kunchang Li, Yali Wang, Yinan He, Yizhuo Li, Yi Wang, Yi Liu, Zun Wang, Jilan Xu, Guo Chen, Ping Luo, et~al.
\newblock Mvbench: A comprehensive multi-modal video understanding benchmark.
\newblock In \emph{Proceedings of the IEEE/CVF Conference on Computer Vision and Pattern Recognition}, pages 22195--22206, 2024{\natexlab{b}}.

\bibitem[Li et~al.(2024{\natexlab{c}})Li, Wang, and Jia]{li2024llama}
Yanwei Li, Chengyao Wang, and Jiaya Jia.
\newblock Llama-vid: An image is worth 2 tokens in large language models.
\newblock In \emph{European Conference on Computer Vision}, pages 323--340. Springer, 2024{\natexlab{c}}.

\bibitem[Lin et~al.(2023)Lin, Ye, Zhu, Cui, Ning, Jin, and Yuan]{lin2023video}
Bin Lin, Yang Ye, Bin Zhu, Jiaxi Cui, Munan Ning, Peng Jin, and Li Yuan.
\newblock Video-llava: Learning united visual representation by alignment before projection.
\newblock \emph{arXiv preprint arXiv:2311.10122}, 2023.

\bibitem[Liu et~al.(2023)Liu, Li, Wu, and Lee]{liu2023visual}
Haotian Liu, Chunyuan Li, Qingyang Wu, and Yong~Jae Lee.
\newblock Visual instruction tuning.
\newblock \emph{Advances in neural information processing systems}, 36:\penalty0 34892--34916, 2023.

\bibitem[Liu et~al.(2024)Liu, Wang, Ma, Wu, Ma, Wei, Jiao, Wu, and Hu]{liu2024kangaroo}
Jiajun Liu, Yibing Wang, Hanghang Ma, Xiaoping Wu, Xiaoqi Ma, Xiaoming Wei, Jianbin Jiao, Enhua Wu, and Jie Hu.
\newblock Kangaroo: A powerful video-language model supporting long-context video input.
\newblock \emph{arXiv preprint arXiv:2408.15542}, 2024.

\bibitem[Maaz et~al.(2023)Maaz, Rasheed, Khan, and Khan]{maaz2023video}
Muhammad Maaz, Hanoona Rasheed, Salman Khan, and Fahad~Shahbaz Khan.
\newblock Video-chatgpt: Towards detailed video understanding via large vision and language models.
\newblock \emph{arXiv preprint arXiv:2306.05424}, 2023.

\bibitem[Maaz et~al.(2024)Maaz, Rasheed, Khan, and Khan]{maaz2024videogpt+}
Muhammad Maaz, Hanoona Rasheed, Salman Khan, and Fahad Khan.
\newblock Videogpt+: Integrating image and video encoders for enhanced video understanding.
\newblock \emph{arXiv preprint arXiv:2406.09418}, 2024.

\bibitem[Monfort et~al.(2021)Monfort, Jin, Liu, Harwath, Feris, Glass, and Oliva]{monfort2021spoken}
Mathew Monfort, SouYoung Jin, Alexander Liu, David Harwath, Rogerio Feris, James Glass, and Aude Oliva.
\newblock Spoken moments: Learning joint audio-visual representations from video descriptions.
\newblock In \emph{Proceedings of the IEEE/CVF Conference on Computer Vision and Pattern Recognition}, pages 14871--14881, 2021.

\bibitem[Ning et~al.(2024)Ning, Zhao, Jin, Ding, and Guo]{ning2024inf}
Zhenyu Ning, Jieru Zhao, Qihao Jin, Wenchao Ding, and Minyi Guo.
\newblock Inf-mllm: Efficient streaming inference of multimodal large language models on a single gpu.
\newblock \emph{arXiv preprint arXiv:2409.09086}, 2024.

\bibitem[Ren et~al.(2023)Ren, Chen, Li, Sun, and Hou]{ren2023testa}
Shuhuai Ren, Sishuo Chen, Shicheng Li, Xu Sun, and Lu Hou.
\newblock Testa: Temporal-spatial token aggregation for long-form video-language understanding.
\newblock \emph{arXiv preprint arXiv:2310.19060}, 2023.

\bibitem[Shang et~al.(2024)Shang, Cai, Xu, Lee, and Yan]{shang2024llava}
Yuzhang Shang, Mu Cai, Bingxin Xu, Yong~Jae Lee, and Yan Yan.
\newblock Llava-prumerge: Adaptive token reduction for efficient large multimodal models.
\newblock \emph{arXiv preprint arXiv:2403.15388}, 2024.

\bibitem[Shu et~al.(2024)Shu, Zhang, Liu, Qin, Zhou, Huang, and Zhao]{shu2024video}
Yan Shu, Peitian Zhang, Zheng Liu, Minghao Qin, Junjie Zhou, Tiejun Huang, and Bo Zhao.
\newblock Video-xl: Extra-long vision language model for hour-scale video understanding.
\newblock \emph{arXiv preprint arXiv:2409.14485}, 2024.

\bibitem[Team et~al.(2024)Team, Georgiev, Lei, Burnell, Bai, Gulati, Tanzer, Vincent, Pan, Wang, et~al.]{team2024gemini}
Gemini Team, Petko Georgiev, Ving~Ian Lei, Ryan Burnell, Libin Bai, Anmol Gulati, Garrett Tanzer, Damien Vincent, Zhufeng Pan, Shibo Wang, et~al.
\newblock Gemini 1.5: Unlocking multimodal understanding across millions of tokens of context.
\newblock \emph{arXiv preprint arXiv:2403.05530}, 2024.

\bibitem[Team(2024)]{team2024internvl2}
OpenGVLab Team.
\newblock Internvl2: Better than the best—expanding performance boundaries of open-source multimodal models with the progressive scaling strategy, 2024.

\bibitem[Touvron et~al.(2023)Touvron, Lavril, Izacard, Martinet, Lachaux, Lacroix, Rozi{\`e}re, Goyal, Hambro, Azhar, et~al.]{touvron2023llama}
Hugo Touvron, Thibaut Lavril, Gautier Izacard, Xavier Martinet, Marie-Anne Lachaux, Timoth{\'e}e Lacroix, Baptiste Rozi{\`e}re, Naman Goyal, Eric Hambro, Faisal Azhar, et~al.
\newblock Llama: Open and efficient foundation language models.
\newblock \emph{arXiv preprint arXiv:2302.13971}, 2023.

\bibitem[Wang et~al.(2024{\natexlab{a}})Wang, Bai, Tan, Wang, Fan, Bai, Chen, Liu, Wang, Ge, et~al.]{wang2024qwen2}
Peng Wang, Shuai Bai, Sinan Tan, Shijie Wang, Zhihao Fan, Jinze Bai, Keqin Chen, Xuejing Liu, Jialin Wang, Wenbin Ge, et~al.
\newblock Qwen2-vl: Enhancing vision-language model's perception of the world at any resolution.
\newblock \emph{arXiv preprint arXiv:2409.12191}, 2024{\natexlab{a}}.

\bibitem[Wang et~al.(2024{\natexlab{b}})Wang, Si, Wu, Zhu, Cao, and Nie]{wang2024retake}
Xiao Wang, Qingyi Si, Jianlong Wu, Shiyu Zhu, Li Cao, and Liqiang Nie.
\newblock Retake: Reducing temporal and knowledge redundancy for long video understanding.
\newblock \emph{arXiv preprint arXiv:2412.20504}, 2024{\natexlab{b}}.

\bibitem[Wang et~al.(2024{\natexlab{c}})Wang, Li, Li, Yu, He, Chen, Pei, Zheng, Wang, Shi, et~al.]{wang2024internvideo2}
Yi Wang, Kunchang Li, Xinhao Li, Jiashuo Yu, Yinan He, Guo Chen, Baoqi Pei, Rongkun Zheng, Zun Wang, Yansong Shi, et~al.
\newblock Internvideo2: Scaling foundation models for multimodal video understanding.
\newblock In \emph{European Conference on Computer Vision}, pages 396--416. Springer, 2024{\natexlab{c}}.

\bibitem[Wei and Chen(2024)]{wei2024visual}
Hongchen Wei and Zhenzhong Chen.
\newblock Visual context window extension: A new perspective for long video understanding.
\newblock \emph{arXiv preprint arXiv:2409.20018}, 2024.

\bibitem[Weng et~al.(2024)Weng, Han, He, Chang, and Zhuang]{weng2024longvlm}
Yuetian Weng, Mingfei Han, Haoyu He, Xiaojun Chang, and Bohan Zhuang.
\newblock Longvlm: Efficient long video understanding via large language models.
\newblock In \emph{European Conference on Computer Vision}, pages 453--470. Springer, 2024.

\bibitem[Wu et~al.(2025)Wu, Li, Chen, and Li]{wu2025longvideobench}
Haoning Wu, Dongxu Li, Bei Chen, and Junnan Li.
\newblock Longvideobench: A benchmark for long-context interleaved video-language understanding.
\newblock \emph{Advances in Neural Information Processing Systems}, 37:\penalty0 28828--28857, 2025.

\bibitem[Xiao et~al.(2023)Xiao, Tian, Chen, Han, and Lewis]{xiao2023efficient}
Guangxuan Xiao, Yuandong Tian, Beidi Chen, Song Han, and Mike Lewis.
\newblock Efficient streaming language models with attention sinks.
\newblock \emph{arXiv preprint arXiv:2309.17453}, 2023.

\bibitem[Xue et~al.(2024)Xue, Chen, Li, Hu, Zhu, Li, Fang, Tang, Yang, Liu, et~al.]{xue2024longvila}
Fuzhao Xue, Yukang Chen, Dacheng Li, Qinghao Hu, Ligeng Zhu, Xiuyu Li, Yunhao Fang, Haotian Tang, Shang Yang, Zhijian Liu, et~al.
\newblock Longvila: Scaling long-context visual language models for long videos.
\newblock \emph{arXiv preprint arXiv:2408.10188}, 2024.

\bibitem[Yang et~al.(2024)Yang, Yang, Zhang, Hui, Zheng, Yu, Li, Liu, Huang, Wei, et~al.]{yang2024qwen2}
An Yang, Baosong Yang, Beichen Zhang, Binyuan Hui, Bo Zheng, Bowen Yu, Chengyuan Li, Dayiheng Liu, Fei Huang, Haoran Wei, et~al.
\newblock Qwen2. 5 technical report.
\newblock \emph{arXiv preprint arXiv:2412.15115}, 2024.

\bibitem[Zhang et~al.(2024{\natexlab{a}})Zhang, Li, Zhang, Pu, Cahyono, Hu, Liu, Zhang, Yang, Li, and Liu]{zhang2024lmmsevalrealitycheckevaluation}
Kaichen Zhang, Bo Li, Peiyuan Zhang, Fanyi Pu, Joshua~Adrian Cahyono, Kairui Hu, Shuai Liu, Yuanhan Zhang, Jingkang Yang, Chunyuan Li, and Ziwei Liu.
\newblock Lmms-eval: Reality check on the evaluation of large multimodal models, 2024{\natexlab{a}}.

\bibitem[Zhang et~al.(2024{\natexlab{b}})Zhang, Zhang, Li, Zeng, Yang, Zhang, Wang, Tan, Li, and Liu]{zhang2024long}
Peiyuan Zhang, Kaichen Zhang, Bo Li, Guangtao Zeng, Jingkang Yang, Yuanhan Zhang, Ziyue Wang, Haoran Tan, Chunyuan Li, and Ziwei Liu.
\newblock Long context transfer from language to vision.
\newblock \emph{arXiv preprint arXiv:2406.16852}, 2024{\natexlab{b}}.

\bibitem[Zhang et~al.(2024{\natexlab{c}})Zhang, Li, Liu, Lee, Gui, Fu, Feng, Liu, and Li]{zhang2024llava}
Y Zhang, B Li, H Liu, Y Lee, L Gui, D Fu, J Feng, Z Liu, and C Li.
\newblock Llava-next: A strong zero-shot video understanding model.
\newblock 2024{\natexlab{c}}.

\bibitem[Zhang et~al.(2024{\natexlab{d}})Zhang, Wu, Li, Li, Ma, Liu, and Li]{zhang2024video}
Yuanhan Zhang, Jinming Wu, Wei Li, Bo Li, Zejun Ma, Ziwei Liu, and Chunyuan Li.
\newblock Video instruction tuning with synthetic data.
\newblock \emph{arXiv preprint arXiv:2410.02713}, 2024{\natexlab{d}}.

\bibitem[Zhou et~al.(2024)Zhou, Shu, Zhao, Wu, Xiao, Yang, Xiong, Zhang, Huang, and Liu]{zhou2024mlvu}
Junjie Zhou, Yan Shu, Bo Zhao, Boya Wu, Shitao Xiao, Xi Yang, Yongping Xiong, Bo Zhang, Tiejun Huang, and Zheng Liu.
\newblock Mlvu: A comprehensive benchmark for multi-task long video understanding.
\newblock \emph{arXiv preprint arXiv:2406.04264}, 2024.

\end{thebibliography}
}

\end{document}